\documentclass{article}
\pdfoutput=1
\usepackage[preprint,nonatbib]{neurips_2024}
\usepackage[utf8]{inputenc} 
\usepackage[T1]{fontenc}    
\usepackage{hyperref}       
\usepackage{url}            
\usepackage{booktabs}       
\usepackage{amsfonts}       
\usepackage{nicefrac}       
\usepackage{microtype}      
\usepackage{xcolor}         
\usepackage{xspace}
\usepackage{graphicx}
\usepackage{algorithm}
\usepackage{algpseudocode}
\usepackage[nointegrals]{wasysym}
\usepackage{bbm}
\usepackage{multirow}
\usepackage{amsmath}
\usepackage{cleveref}

\newcommand{\name}{ElastiFormer\xspace}

\title{\name: Learned Redundancy Reduction in Transformer via Self-Distillation}

\author{%
  Junzhang Liu$^\ast$ \\
  Department of Computer Science\\
  Columbia University\\
  \And 
  Tingkai Liu$^{\ast,\dagger}$\thanks{Equal contributions}\thanks{Corresponding: \texttt{tiliu@cshl.edu, stephen.xia@northwestern.edu}} \\
  NeuroAI Scholars\\
  Cold Spring Harbor Laboratory\\
  \And 
  Yueyuan Sui\\
  Electrical and Computer Engineering \\
  Northwestern University\\
  \And 
  Stephen Xia$^\dagger$ \\
  Electrical and Computer Engineering \\
  Northwestern University\\
}

\begin{document}

\maketitle

\begin{abstract}
We introduce \name, a post-training technique that adapts pretrained Transformer models into an elastic counterpart with variable inference time compute. \name introduces small routing modules (as low as .00006\% additional trainable parameters) to dynamically selects subsets of network parameters and input tokens to be processed by each layer of the pretrained network in an input-dependent manner. The routing modules are trained using self-distillation losses to minimize the differences between the output of the pretrained-model and their elastic counterparts. As \name makes no assumption regarding the modality of the pretrained Transformer model, it can be readily applied to all modalities covering causal language modeling, image modeling as well as visual-language modeling tasks. We show that 20\% to 50\% compute saving could be achieved for different components of the transformer layer, which could be further reduced by adding very low rank LoRA weights (rank 1) trained via the same distillation objective. Finally, by comparing routing trained on different subsets of ImageNet, we show that \name is robust against the training domain. 
\end{abstract}
\section{Introduction}
\label{sec:intro}

The success of Transformer \cite{transformer} models across various domains has led to increasingly large and computationally intensive architectures. While these models achieve impressive performance, recent works have demonstrated significant redundancies in both the parameters and computations in pretrained large transformer architectures - which can either manifest as unnecessary processing of tokens~\cite{bolya2022token,rao2021dynamicvit}, or activation~\cite{dalvi2020analyzing} of model parameters that contribute minimally to the final output. 

Several recent studies have attempted to locate these redundancies~\cite{dalvi2020analyzing,chen2022principle,pan2021ia} and to leverage such redundancies via modified/optimized model architectures (e.g. Mixture-of-Depth which skips tokens around transformer layers \cite{raposo2024mixture}). However, such approaches often require architectural modifications, extensive retraining or even training from scratch (such as commonly done for sparse Mixture-of-Expert models \cite{fedus2022switchtransformersscalingtrillion,jiang2024mixtral}).

We address these challenges by introducing {\bf \name}, a post-training technique that transforms pre-trained Transformer models into elastic alternatives with variable inference time compute. \name introduces lightweight routing modules (as little as $0.0006\%$ additional parameters) that dynamically select subsets of network parameters and input tokens to be processed in an \emph{input-dependent} manner. These routing modules are trained using self-distillation loss to minimize differences between the output of the pretrained model and their elastic counterparts. Crucially, \name makes no assumptions about the modality of the pretrained Transformer model, allowing it to be effectively applied across language, image, and visual-language modalities.

Through extensive experimentation, we demonstrate that 38\% (Mutli-Head Attention, MHA) and 56\% (Multi-Layer Perceptron, MLP) active parameters are required for \name to match the performance of the base pretrained model in the language domain. We also show that $\sim20\%$ tokens can be dropped from MLP processing (routed to output via residual without going through MLP), while almost all tokens need to be processed by MHA. However, by adding Low-Rank Adaptation (LoRA) weights to MHA with as low as rank 1 learnable parameters, we can drop 20\% tokens from MHA processing as well. We observe similar redundancy in parameters and token processing in the visual domain and show that applying \name to \emph{even} layers of the pretrained ViT backbone significantly improves performance at equivalent level of compute to applying \name to all layers. Applying \name to visual-language models, we show that 40\% of image tokens can be dropped before being decoded by the language decoder without significantly impacting performance. Finally, by comparing routing modules in Elasti-ViT trained using different subsets of ImageNet, we show that the learned routing is robust to changing data distributions.

\begin{figure*}[t]
    \centering
    \includegraphics[width=.7\linewidth]{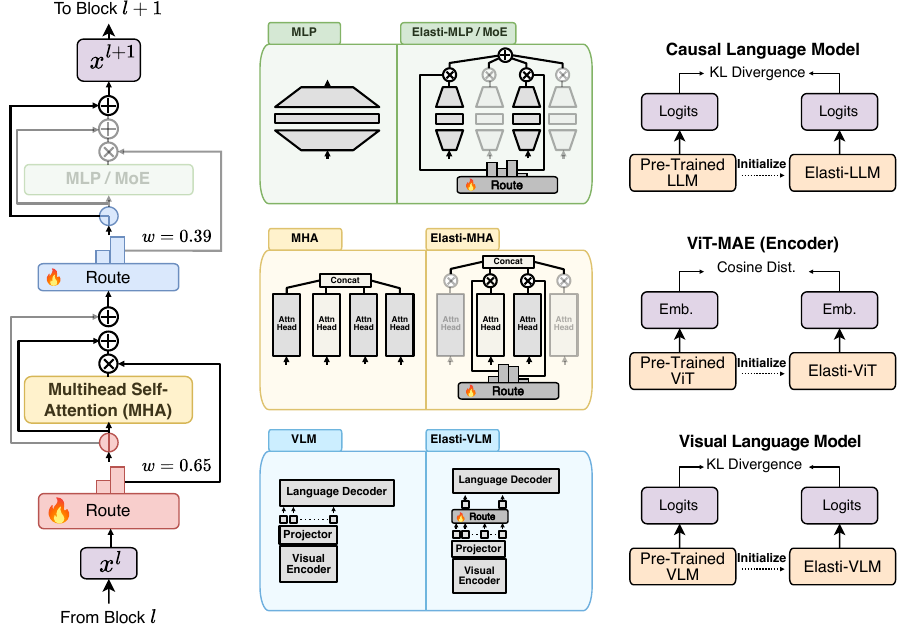}
    \caption{
        Overview of \name for language, visual, and multi-modal transformers. 
        (Left) Illustration of learned routing modules around Multi-Head Attention (MHA) and Multi-layer Perceptron (MLP) modules of a pretrained transformer model.
        (Middle) Illustration of learned routing modules inside MLP and MHA modules, and learned routing that selects a subset of image tokens that provide multi-modal input to language decoder in VLMs.
        (Right) Illustration of self-distillation training objectives across modalities. Note that for Visual Transformers (ViT), the example provided here is from Masked Auto Encoding (MAE) ViT.
    }
    \label{fig:fig1}
\end{figure*}

In summary, our main contributions are:
\begin{enumerate}
    \item We introduce \name, a post-training technique that converts pretrained transformer models across language, visual, and multi-modal modalities into flexible alternatives with variable inference time compute. \name is compatible with other efficient post-training techniques.
    \item We demonstrate that with less than $0.3\%$ (as low as $0.00006\%$) additional learnable parameters, \name can reduce the total active parameters or active tokens by 20\% to 50\% for all modalities without noticeable impact on performance. We also provide extensive analysis of the degree of redundancy in different components of the transformer architecture across modalities.
    \item We show that the learned routing is robust to different distributions of training data, thereby providing a reliable representation for interpreting learned representations in pretrained models.
\end{enumerate}

\section{Related Works}

\noindent
\textbf{Efficient Transformer Processing} There is a large body of work aimed at improving transformer efficiency that falls into several categories: mixture of experts (MoE), mixture of depths (MoD), pruning, and quantization~\cite{xiao2023smoothquant,shao2023omniquant,dettmers2024qlora,liu2021post,lin2021fq,liu2023oscillation}. Our work is compatible with existing post-training techniques, such as QLoRA~\cite{dettmers2024qlora}, but is most similar in spirit to MoE, MoD, and pruning. Since the introduction of MoE in transformer architectures~\cite{shazeer2017outrageously}, there have been a flurry of developments, including more efficient MoE architectures~\cite{jain2024mixture,dai2024deepseekmoe}, sparsification of dense MLPs~\cite{zhu2024llama}, incorporating gating functions for routing~\cite{zhang2021moefication,jiang2024mixtral}, and realizing LLMs at scales larger than ever~\cite{fedus2022switchtransformersscalingtrillion}.

Pruning is another class of techniques used to statically remove redundant weights (unstructured)~\cite{chen2020lottery,huang2021sparse} or smaller blocks (structured) in the MHA or FFN layers~\cite{ma2023llm,fang2023depgraph,kim2024shortened,an2024fluctuation,xia2023sheared,xia2022structured,chavan2022vision,yu2023x}. Other works focus on pruning or merging context in the KV cache~\cite{anagnostidis2024dynamic,xu2024think} or individual patches in vision transformers~\cite{tang2022patch,jian2024expedited}.

Most similar to our work is the recent development of MoD~\cite{raposo2024mixture}, which proposes learnable routers that ``skip'' less relevant transformer layers in an input-dependent and compute-adjustable manner. In contrast to ``early exit'' methods~\cite{schuster2022confident,liu2021anytime}, which skip any remaining transformer layers, MoD may skip middle layers and still process later layers. Other works have shown varying levels of spatial and temporal redundancies in transformer architectures~\cite{dutson2023eventful,chen2022principle,dalvi2020analyzing}, and introduce methods for efficiently identifying and removing them during inference (e.g., skipping attention heads~\cite{he2024matters}). In contrast, our work provides a general learning framework for discovering and bypassing redundancies to enable more efficient processing pathways through pretrained transformer models.

\noindent
\textbf{Knowledge Distillation (KD) and Domain Generalization.} KD is a class of techniques that transfer knowledge from one model to another (often smaller), where the structure of the models are generally fixed~\cite{hinton2015distilling,sun2019patient,jiao2019tinybert,wang2020minilm,sanh2019distilbert,touvron2021training,hao2022learning,sun2020mobilebert,wu2022tinyvit,lin2022knowledge,yu2022unified}. Self-distillation (SD) is a sub-class of KD, where the student and teacher models are the same architecture. In many cases, the SD aims to distill knowledge learned in deeper layers to shallow layers in the same model, which has shown to reduce overfitting, improve domain generalization, and embed semantic information that is not commonly seen in non-SD models~\cite{sultana2022self,park2022self,caron2021emerging,suh2023tasked,jiang2022self,ma2023transformer,jeoung2023self,lee2022self,wang2022last}. Most of these works focus on pretraining to improve task performance, while we incorporate SD as a mechanism to self learn router weights to reduce computation in an input-dependent and non-sequential manner.

\noindent
\textbf{Mechanistic Interpretability (MI).} MI has recently emerged as a promising class of techniques for interpreting deep neural networks and transformer architectures~\cite{olah2020zoom,rai2024practical} by decomposing models into smaller components and identifying human interpretable ``features'' and their pathways that form subnetworks or ``circuits''. A key question in MI is to explore the notion of ``universality'' or how similar features, activations, or circuits arise between different models and tasks. While many works have explored and identified how each component (e.g., MHA and FFN sublayers) functions within transformer architectures~\cite{elhage2021mathematical,mcdougall2023copy,meng2022locating,stolfo2023mechanistic,geva2020transformer}, there are mixed results when it comes to discovering universality. Several works have identified similar components, such as induction and duplication heads, that develop across different models and tasks~\cite{olsson2022context,gould2023successor,wang2022interpretability,merullo2023circuit}, while others observe that different weight initializations for the same task result in different circuits and representations~\cite{zhong2024clock,chughtai2023toy}. We believe our work could be used to study the interpretability of transformer architectures by analyzing the learned routers across different model architectures and domains.
\section{Redundancy in Transformer Architecture}\label{sec:redundancy}

\begin{figure}[t]
    \centering
    \includegraphics[width=.7\linewidth]{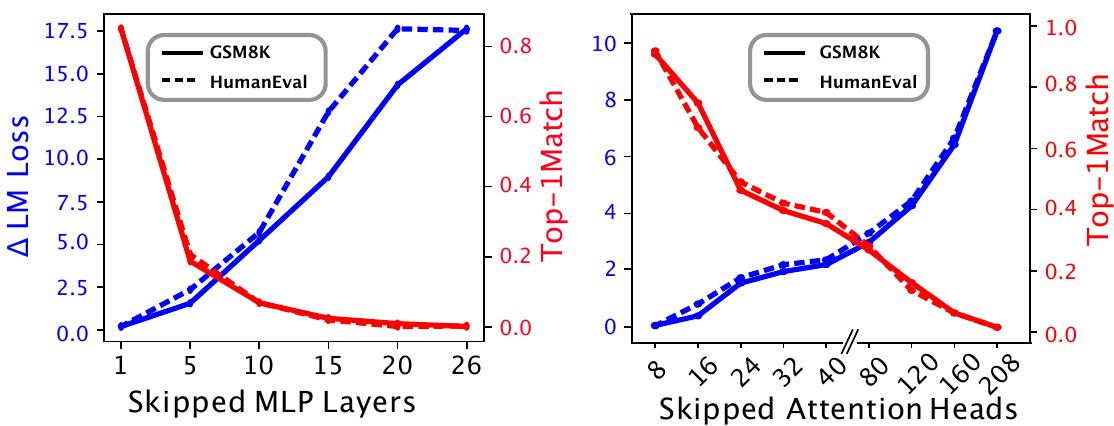}
    \caption{Difference in language modeling loss (blue) and top-1 token prediction agreement (red) between pretrained Gemma2 model and Gemma2 model with skipped MLP layers (left) or attention heads in MHA (right).
    Experiments are performed for both GSM8K (solid line) and HumanEval (dashed line) datasets.
    }
    \label{fig:redundancy_analysis}
\end{figure}

We first demonstrate that pretrained transformer models have data-dependent redundancies in both their multi-head attention (MHA) and multi-layer perceptron (MLP) modules. We progressively dropped attention heads and MLP layers in a pretrained \texttt{Gemma-2-2b-it} \cite{gemma2} model, and quantified the performance degredation on both mathematical reasoning (GSM8K\cite{cobbe2021gsm8k}) and code generation (HumanEval\cite{humaneval_dataset}) tasks. Note that in these experiments, no additional learnable parameters were introduced. 

For each experiment, we randomly select a progressively larger number of attention-heads and MLP layers to remove and calculate the total differences in language modeling losses as well as the percentage agreement of the vocabulary index predicted by the base and pruned model. As illustrated in Figure \ref{fig:redundancy_analysis}, skipping a small number of heads or layers has negligible impact on model performance, with faster performance deterioration observed when skipping more MLP layers than removing attention heads. Importantly, the performance scaling differ between GSM8K and HumanEval datasets, indicating that the redundancy in the pretrained LLM is data-dependent. These results motivated us to explore learned, data-dependent routing modules that not only skip MLP layers and attention heads in a learnable manner, but also skip attention layers and subsets of MLP weights. As later demonstrated, \name offers significantly higher degree of compute savings over static pruning at comparable performance degradation.

\section{\name ~- Learned Routing of Transformer Models via Self-Distillation}

Given the redundancy in pre-trained Transformer models, we seek a lightweight post-training technique to minimize the number of active parameters necessary without significantly deteriorating overall performance. Consequently, we propose \name which introduces lightweight learned, input-dependent routing modules that route input tokens (regardless of modality) through a subset of the pre-trained Transformer network.

\begin{figure}[t]
    \centering
    \includegraphics[width=0.7\linewidth]{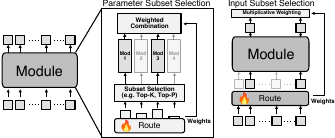}
    \caption{
    Illustration of the two subset selection schemes employed in the current work.
    For parameter subset selection, the sub-modules (Mod 1-4) that are selected by the routing scheme can either refer to the attention heads in MHA or experts in MLP. Note that to create experts in MLP from a pretrained dense MLP layer, we first transform the dense MLP parameters to block matrices that form the experts. In either routing scheme, the routing weights are multiplied
    with the output to ensure gradient flow.
    }
    \label{fig:routing}
\end{figure}

Depending on the modality, \name introduces 4-5 routing modules that control routing around and within all modules of the transformer architecture (i.e. Multi-Head Attention, MLP). The routing schemes can be roughly divided into two categories (see Figure.~\ref{fig:routing}):
\begin{itemize}
    \item Input Subset Selection: For this type of subset selection, given a sequence of $T$ tokens, we select $k$ to be processed by a given module (e.g. MHA, MLP). This routing scheme saves computational cost by reducing the total number of \emph{input tokens} processed. Examples of routing in \name that employ this scheme are routing around MHA, around MLP (Figure~\ref{fig:fig1}(Left)), and visual token selection in VLM (Figure~\ref{fig:fig1}(Mid-Bottom)).
    \item Parameter Subset Selection: For this type of subset selection, the total number of inputs to a given module remains unchanged. Instead, we save computational cost by reducing the number of active parameters within a given module that is used to process the given input. Examples of routing in \name that employ this scheme are routing within MHA (attention head selection) and within MLP (expert selection). 
\end{itemize}

\subsection{Parameter Subset Selection}
We consider the routing problem where we are given input token $x_t \in \mathbb{R}^D$, and $M$ subsets of sub-networks $f_1,\ldots,f_M$ ($M \ll D$). A simple router (linear projector) is trained to select top $k, 1\le k \le M$ sub-networks to process the given input following the procedure outlined in Algorithm~\ref{alg:param_linear}. 

\begin{algorithm}
\caption{Parameter Subset Selection w/ Linear Router}
\begin{algorithmic}[1]
\Require Input token $x_t \in \mathbb{R}^D$
\Require Set of sub-networks $\{f_m\}_{m=1}^M$
\Require Number of routes to select $k$, where $1 \leq k \leq M$
\Require Router parameters $W_r \in \mathbb{R}^{M \times D}$

\State $w_t \gets  M\cdot \text{softmax}(W_r x_t)$ \Comment{Routing Weights}

\State $\{(m_i, w_{t,m_i})\}_{i=1}^k \gets \text{top-k}(w_t, k)$ \Comment{Subset Selection}

\For{$i = 1$ to $k$}
    \State $y_{t, m_i} \gets w_{t,m_i} \cdot f_{m_i}(x_t)$ \Comment{Sub-network Forward}
\EndFor

\Return $y_t = \text{Aggregate}(\{y_{t,m_i}\}_i)$
\end{algorithmic}
\label{alg:param_linear}
\end{algorithm}

At the high level, the parameter subset selection works essentially the same as the Mixture-of-Expert routing with a straight-through estimator for gradient propagation, with the noticeable difference that the weight vector is normalized to sum to $M$. This additional normalization ensures that when the router selects all sub-networks (number of routes $k=M$) with uniform routing weights $w_{t,i} = 1 ~ \forall i=1,\ldots,k$, the overall input/output of the routed network is \emph{exactly} the same as the pre-trained model without routing.

The sub-network $\{f_m\}_m$ could either refer to processing via individual attention heads or via an expert in the Mixture-of-Expert MLP module. As many pre-trained transformer models have dense MLP layers (no experts), \name converts a dense MLP module to a MoE counterpart losslessly by breaking parameters into block matrices. Consider a simple MLP with 1 hidden layer, we can rewrite the input/output relationship of the dense MLP as the equivalent MoE MLP with 2-experts:
\begin{equation*}
y_t = W_2 ~ \sigma(W_1 x_t) =
\begin{bmatrix} W_{2,1} & W_{2,2} \end{bmatrix} \sigma\left(\begin{bmatrix} W_{1,1} \\ W_{1,2} \end{bmatrix} x_t\right),
\end{equation*}
where $\sigma(\cdot)$ is some element-wise nonlinearity, and $W_{1,i}, W_{2,i}$ are block matrices obtained by splitting the weight matrices row-wise and column-wise respectively. While we found decomposing the pretrained weights to block matrices sufficient, we acknowledge that alternative initialization of MoE MLP from dense MLP exists and have been explored by contemporary works \cite{zhu2024llama,zhang2021moefication,zuo2022moebert}. We leave further explorations of MoE initialization from dense MLP in the context of \name for further work.

\paragraph{Input Subset Selection} 

We consider the routing problem where we are given input token $x_{1:T} \in \mathbb{R}^{T\times D}$ and a given module $f:\mathbb{R}^D\mapsto \mathbb{R}^D$. A router is trained to select top $k, 1\le k \le T$ tokens to process by the module following the Algorithm~\ref{alg:token_linear}. 

\begin{algorithm}
\caption{Input Subset Selection w/ Linear Router}
\begin{algorithmic}[1]
\Require Input tokens $x_{1:T} \in \mathbb{R}^{T\times D}$
\Require Module $f$
\Require Number of tokens to select $k$, where $1 \leq k \leq T$
\Require Router parameters $W_r \in \mathbb{R}^D$

\State $w \gets  \text{softmax}(W_r x_{1:T})$ \Comment{Routing weights $w\in\mathbb{R}^{T}$}

\State $I \gets \text{top-k}(w, k)$ \Comment{Index set of selected input }

\State $y_{1:T} \gets {\bf 0} \in \mathbb{R}^{T\times D}$ \Comment{Zero-initialized output}
\State $y_{t\in I} \gets w_{t\in I} \cdot f(x_{t\in I})$

\Return $y$
\end{algorithmic}
\label{alg:token_linear}
\end{algorithm}

As opposed to parameter subset selection where the router outputs a $M$-dimensional logits for $M$ sub-networks, the routers in input subset selection output scalar-valued logits for each input token. The top-k input tokens are then processed by the subsequent module (MHA or MLP), and the output is added to a zero-initialized output tensor with the same shape as the un-selected input $x_{1:T}$. 

Note that, as discussed in \cite{raposo2024mixture}, for causal language models that generate tokens auto-regressively, the top-k token selection only applies to the training phase and not the inference phase - where causality dictates whether a given token will be in the top-k cannot be determined without first completing the generation of the entire sequence. Authors of \cite{raposo2024mixture} proposed 
two approaches to predict whether a given token will be in the top-k during inference time based on 1) the output logits of the same router that computes the routing weights, or 2) output logits of an additional dedicated classification MLP that similarly maps input embedding $x_t$ to a scalar logit value. 
Within the context of \name, we found the two methods of top-k prediction during inference to give similar performance and opted to use the former (same router for top-k prediction) to minimize additional learnable parameters.

\subsection{Objective Function}\label{sec:objective}
\begin{figure}[t]
    \centering
    \includegraphics[width=\linewidth]{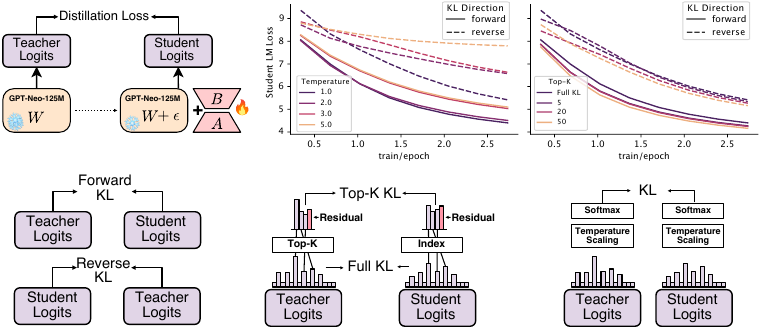}
    \caption{
        Comparison between different distillation losses for language output modality.
        The three types of variations of the KL-divergence objectives are illustrated on the bottom row.
    }
    \label{fig:distill}
\end{figure}

The primary training objective of \name is distillation loss $\mathcal{L}_{\text{distill}}$. For causal language modeling and visual-language modeling where the output modality is language, we compute KL-divergence between the student model's output probability $p_\text{student}$ and the teacher model's output probability $p_\text{teacher}$. We compared different modifications of the KL-divergence objective for distillation losses on a toy problem, where the teacher model is a pretrained \texttt{GPT-Neo-125M} \cite{gpt-neo} model, and the student model is initialized from teacher model with Gaussian parameter noised and a rank-32 LoRA adapter for \texttt{q\_proj},\texttt{v\_proj} components in the MHA module. The student model is trained on the GSM8K training dataset using combinations of three different types of modifications to the KL-divergence objective (see also Figure~\ref{fig:distill}):
\begin{itemize}
    \item Forward $D_{KL}(p_\text{student}||p_\text{teacher})$ versus Reverse KL $D_{KL}(p_\text{teacher}||p_\text{student})$ \cite{reverse_kl_1, reverse_kl_2}
    \item Top-K KL \cite{context_distill}: convert the teacher model's output probability over vocabulary to a $k+1$ dimensional vector, where the first $k$ values correspond to the probability of the top-k tokens in the vocabulary with the highest probability, and the last value is the residual probability to ensure the $k+1$ probability vector sums to 1. The student model's output is arranged using the top-k token indices of the teacher model.
    \item Temperature scaling \cite{temp_scale_distill}: divide the logits of student and teacher model outputs by a positive scalar temperature before converting to probability distribution via softmax.
\end{itemize}
As shown in Figure~\ref{fig:distill}, the language modeling loss of the student model on the evaluation dataset shows that forward KL loss of top-50 tokens leads to the best performance and fastest convergence rate, which is the distillation loss we adopted in the current work for both language modeling and visual-language modeling tasks.
For image modeling (applying \name to ViT-MAE \cite{vit_mae}), we chose the cosine distance between the student image encoder's output token embedding and that of the teacher model for simplicity. Note that as the output token embedding of the ViT-encoder is normalized via LayerNorm, minimizing cosine distance is equivalent to maximizing the inner product between student and teacher token embedding. 

In addition to the distillation loss, \name training involves additional auxiliary losses for the two types of routing modules described above.

For routing modules that perform \emph{parameter} subset selection, the routers are additionally trained using load-balancing loss $\mathcal{L}_\text{load}$ which minimizes the weighted sum of input tokens processed by each sub-network. While such load-balancing loss is crucial for MoE models where both the experts and the routers are trainable to prevent mode collapse and ensure balanced sub-network utilization, its effect is much less pronounced for \name where the sub-networks are frozen. Nevertheless, we found this auxiliary loss to be beneficial for stabilizing training of router modules especially for very low $k$ values (very few sub-networks activated). 

For routing modules that perform input subset selection, we only include auxiliary loss in the case of \emph{causal} language modeling. As mentioned in the previous section, during inference, scalar logits of a token selection module are used to determine if a given token will likely be in the top-k of the final generated sequence or not. Consequently, during the training phase, binary cross-entropy loss $\mathcal{L}_{\text{top-k}}$ between the scalar logits and one-hot encoding of whether the corresponding token is indeed in top-k is introduced as an auxiliary loss to the input subset selection for causal LMs \cite{raposo2024mixture}.

As such the overall objective for \name is:
\begin{equation}
    \mathcal{L} = \mathcal{L}_{\text{distill}} + \lambda_{\text{load}} \mathcal{L}_{\text{load}} + \underbrace{\lambda_{\text{top-k}} \mathcal{L}_{\text{top-k}}}_\text{for Causal LM}
\end{equation}
In practice, we found $ \lambda_{\text{load}} = \lambda_{\text{top-k}} = 1$ to be sufficient for convergence of all evaluated model architectures. 
\section{Experiments}
In this section, we describe in detail the experimental setup and findings of the current work.
For reference, we included the total number of additional parameters introduced for each \name experiment in Table~\ref{tab:param} which highlights that \name introduces \emph{minimal} ($0.25\% \sim 0.00006\%$) additional learnable parameters while achieving significant redundancy reduction across modalities. Unless otherwise specified, all experiments were performed on 1 NVIDIA H100 NVL 96GB GPU with AdamW optimization of learning rate \texttt{1e-4} and cosine learning rate scheduler with 3\% warmup.

\begin{table*}[t]
\centering
\resizebox{.7\linewidth}{!}{
\begin{tabular}{@{}lllllll@{}}

                                  &              &                           & \multicolumn{2}{c}{Elasti-LLM} & Elasti-ViT    & Elasti-VLM     \\ \cmidrule(l){4-7} 
Selection                              & Module       & \# Params.                & Gemma2-2B      & Phi3-Mini     & ViT-MAE-L     & LLaVA-1.5   \\ \midrule
\multirow{4}{*}{Input}  & MLP          & $L\times(D+2)$                   & 60K (.003\%)  & 98K (.003\%) & 25K (.01\%)  & -              \\
                                  & MHA          & $L\times(D+2)$                   & 60K (.003\%)  & 98K (.003\%) & 25K (.01\%)  & -              \\
                                  & VLM/L & $D+2$                       & -              & -             & -             & 4K (.00006\%) \\
                                  & VLM/M    & $D^2+2D+2$ & -              & -             & -             & 17M (.23\%)   \\ \midrule
\multirow{2}{*}{Param.} & MLP          & $L\times(D \times M)$                 & 2M (.08\%)    & 3M (.08\%)   & 0.8M (.24\%) & -              \\
                                  & MHA          & $L\times(D \times M)$                 & 2M (.08\%)    & 3M (.08\%)   & 0.8M (.24\%) & -              \\ \bottomrule
\end{tabular}
}
\caption{Number of trainable parameters introduced by \name with percentage of total number of parameters of pre-trained base model shown in parenthesis. $L$ is the number of layers, $D$ is the hidden dimension and $M$ is the number of sub-networks (number of attention heads or number of experts in MoE-MLP). VLM/L,VLM/M refer to linear or MLP routing modules applied to image tokens selection in VLM. }
\label{tab:param}
\end{table*}

\subsection{Elasti-LLM - \name for Causal Language Model} \label{sec:elasti-llm}
\paragraph{Experimental Setup}
We applied \name to both \texttt{Phi-3.5-mini-instruct} \cite{phi3} and \texttt{Gemma-2-2b-it} \cite{gemma2} pretrained language models, where
the models were trained via self-distillation on the GSM8K \cite{cobbe2021gsm8k} training set (7.5K question/answer pairs) for 3 epochs and batch-size of 32. 

\begin{figure}[t]
    \centering
    \includegraphics[width=\linewidth]{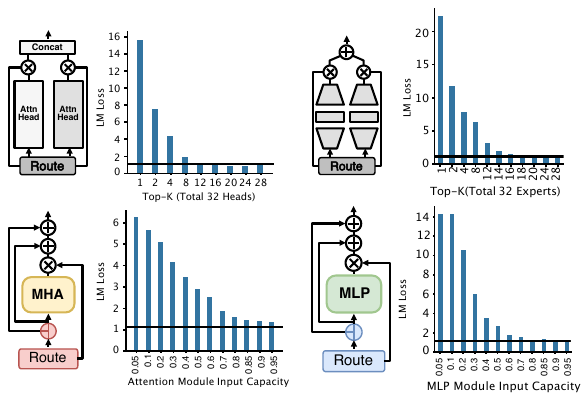}
    \caption{Scaling of various modules in Elasti-LLM against compute for elastic \texttt{Phi-3.5-mini-instruct}. LM Loss of the pretrained teacher model is shown as the horizontal black line in each of the subfigures.}
    \label{fig:phi3_gsm8k_train}
\end{figure}

\paragraph{Scaling of Performance vs. Capacity}
We begin by performing an extensive ablation study of the 4 types of routing in Elasti-LLM shown in Figure~\ref{fig:phi3_gsm8k_train}.

As shown in the figure, we are able to achieve the same level of performance (as measured by LM Loss) as the teacher model with 38\% attention heads (12/32), 56\% active parameters in MLP (18/32 experts initialized from the pretrained dense MLP module), 80\% token processing through the MLP module. We note that deactivating 20/32 attention heads in \emph{each} layer corresponds to a total number of 640 skipped attention heads, which is dramatically higher than the pruning results shown in Figure~\ref{fig:redundancy_analysis}.

\begin{figure}[t] 
    \centering 
    \includegraphics[width=.8\linewidth]{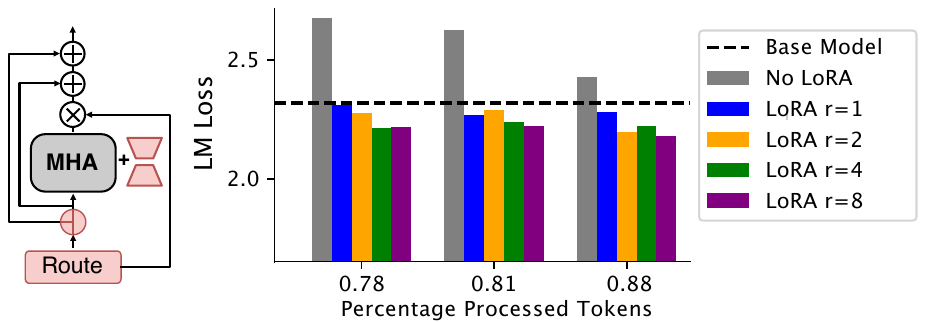} 
    \caption{
        Scaling of \texttt{Gemma-2-2B-it} across a percentage of processed tokens with LoRA adapter for Q, V projections in MHA. The \name module is trained with input subset selection for both MHA and MLP modules, as well as parameter subset selection for the MLP module (with 4 experts and top-2 selection).
    }
    \label{fig:capacity} 
\end{figure}

In contrast to the above-mentioned 3 routing schemes, input selection for the \emph{attention} module could not reach the base model performance even with high input capacity. This is likely due to the context-dependency of processing in the MHA module, which makes input selection that is only based on single token embedding (without context) challenging. While this result appears contradictory to those reported in MoD \cite{raposo2024mixture}
, we note that a key difference between \name and MoD is that modules in \name are completely frozen apart from lightweight routing modules. In contrast, the MoD model trains both the model backbone and routing module simultaneously. Motivated by this, we experimented with adding a small amount of learnable parameters to the MHA module to compensate for the deterioration of performance caused by input selection. We added LoRA modules to the MHA's \texttt{q\_proj},\texttt{v\_proj} weights, which allowed us to achieve the same performance of the teacher model with relatively low lora-rank. The results are shown in Figure~\ref{fig:capacity}, which compares the impact of varying LoRA ranks (r) for an elastic \texttt{gemma-2-2b-it} model across different capacity levels. Even with rank 1 adapters added to Q,V projections in MHA modules ($0.008\%$ trainable parameters), LoRA-enhanced Elasti-LLM achieves performance comparable to the teacher model, with additional trainable weights (higher ranks) leading to even lower loss. Interestingly, we observe that the LM loss of Elasti-LLM can even be lower than that of the base pretrained model, which is consistent with previous literature on self-distillation improving model performance~\cite{sultana2022self,caron2021emerging}.

\subsection{Elasti-ViT - \name for Vision Transformer}
\paragraph{Experimental Setup}
We applied \name to \texttt{vit-mae-large} \cite{vit_mae} pretrained ViT models and trained the \emph{encoder} part of the ViT-MAE model via self-distillation on a 10\% subset of the ImageNet-1K \cite{imagenet} training set ($\sim$1M images) for 3 epochs with batch-size of 900. The training objective minimizes the cosine distance between Elasti-ViT's output token embedding and that of the pretrained ViT-MAE model, while the evaluation is done by comparing the cosine similarities between the \emph{decoder} output between student and teacher models (reconstructed images). 

\paragraph{Scaling of Performance vs. Capacity}
Similar to scaling analysis for causal language models, we performed scaling analysis of Elasti-ViT for different level of subset selection of the 4 different types of routing modules. We evaluate the performance of each experiment by cosine similarity between the MAE decoder (pretrained and frozen) outputs when given encoder output of Elasti-ViT encoder and the based encoder. Example reconstructions are shown in the Supplementary Materials. 

\begin{figure}[t]
    \centering
    \includegraphics[width=\linewidth]{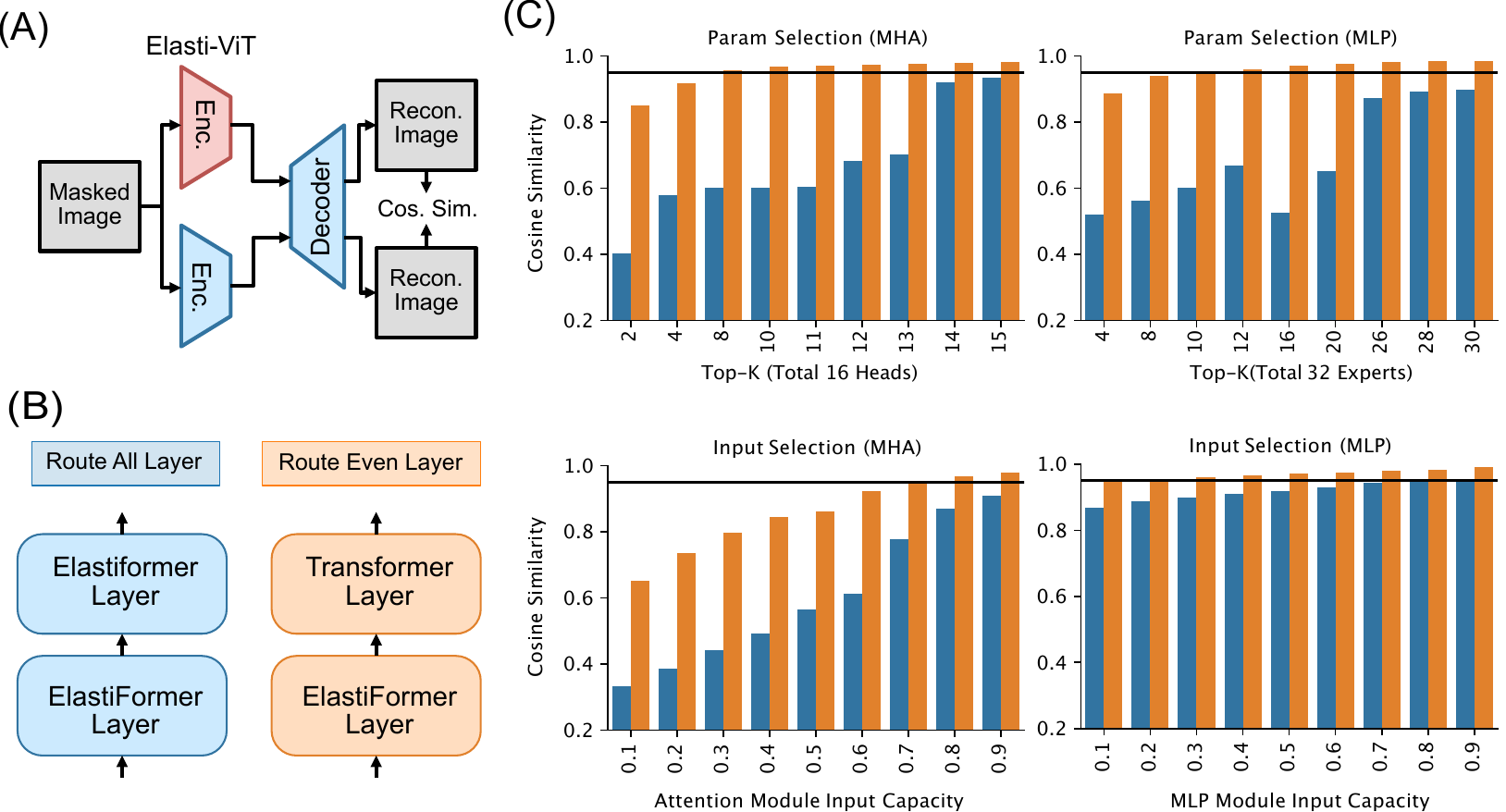}
    \caption{
        Scaling of Elasti-ViT (\texttt{ViT-MAE-L}) performance on held-out ImageNet-1K test set as measured by cosine similarity between MAE decoder output when given encoder output of Elasti-ViT and base pretrained encoder (as shown in (A)). Experiments are shown for both ViT-MAE with \name for all layers or only even layer (as shown in (B)). Horizontal black lines in (C) indicate a cosine similarity of 0.95, which we use as a threshold for when the Elasti-ViT recovers the output of the pretrained ViT-MAE model. Note that as shown in (A), only the encoder has a trainable router in the Elasti-Vit experiments.
    }
    \label{fig:vit}
\end{figure}

Contrary to the experiments in language modeling, results in Figure~\ref{fig:vit} show that only input selection for MLP module is able to achieve $>0.95$ cosine similarity with teacher model's decoder output with input capacity $>0.8$ (dropping 20\% input tokens to MLP modules). This is potentially caused by the much smaller size of \texttt{ViT-MAE-L} models, which has 330M parameters which less than 10\% of the size of \texttt{Phi-3.5-mini-instruct}. 

We explore an alternative technique to improve the performance of Elasti-ViT, instead of adding learnable LoRA modules, we applied ElastiFormer to only \emph{even} layers of the ViT-MAE model. Applying ElastiFormer to even layers reduces computational savings by half while simultaneously reducing the learnable routing parameters by half. As shown in Figure~\ref{fig:vit}, when Elasti-ViT with even layer routing significantly improves the model performance, where 50\% MHA Attention-Heads, 31\% MLP active parameters, 70\% active tokens for MHA and 10\% active tokens for MLP are sufficient to achieve $>0.95$ cosine similarity to the teacher model's decoder output. 
More importantly, by preserving computing in half of the layers, we are able to achieve higher saturing performance of Elasti-ViT. Note that this approach to applying \name to even layers is also compatible with adding LoRA weights, which could further improve computational savings as in the case for Elasti-LLM.

\begin{figure}[t]
    \centering
    \includegraphics[width=0.5\linewidth]{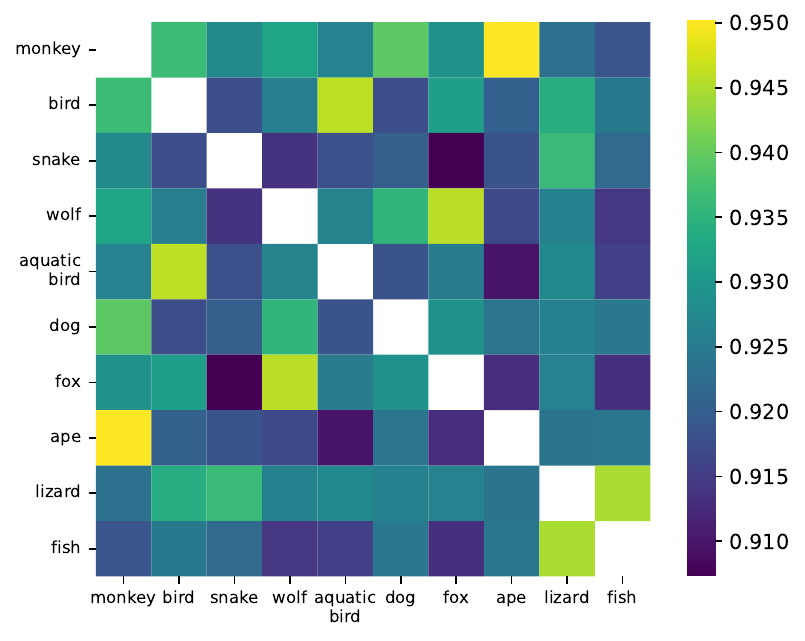}%
    \includegraphics[width=0.5\linewidth]{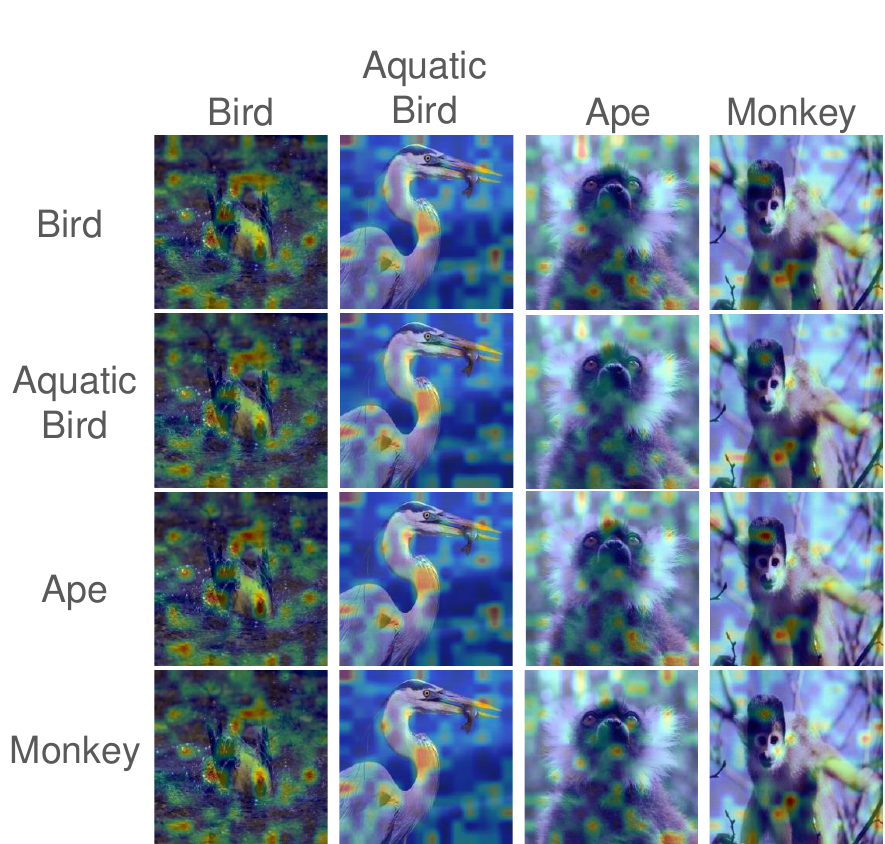}
    \caption{Comparing learned routing of Elasti-ViT trained using different subsets of ImageNet.
    (Left) Cosine similarities between router logits for ten classes in ImageNet. Labels indicate the class of images that an Elasti-ViT instance is trained on. (Right) Heatmap of patches selected by different instances of Elasti-ViT routers. Each row corresponds to a different Elasti-ViT instance trained on a given image class.}
    \label{fig:logit_sim}
\end{figure}

Additionally, we examined the robustness of learned routing of Elasti-ViT against different distributions of training data. To this end, we trained 10 instances of Elasti-ViT using 10 subsets of ImageNet based on categories proposed by \cite{clustered_imagenet_labels}. We then computed the pairwise cosine similarities between router activations of the 10 Elasti-ViTs instances on held-out evaluation images, as shown by the 10x10 similarity matrix shown in \Cref{fig:logit_sim}(Left). All routers shown high similarities, with routers trained on similar classes (e.g. ape and monkey, aquatic bird and bird) show higher similarities than others. The robustness of the learned routing to training data is further demonstrated in \Cref{fig:logit_sim} (Right), where similar the patches of the same image are selected by Elasti-ViT routers trained on different image classes.

\subsection{Elasti-VLM - \name for Visual Language Model}
\paragraph{Training and Evaluation Setup}
We applied \name to \texttt{llava-v1.5-7b} \cite{llava_paper} pretrained visual-language models, which
the was trained via self-distillation on a 10K subset of the 665K \texttt{LLaVA-Instruct} dataset for 1 epoch with batch-size of 32.
 
\paragraph{Scaling of Performance vs. Capacity}
Since experiments for causal LMs and ViTs have established results on redundancy in image encoder and language decoder respectively, we focus our experiments on Elasti-VLM to redundancy in image tokens (output of the visual encoder) processed by the language decoder. In particular, we added an input subset selection routing module that selects the top-k image tokens to be processed by the language decoder as shown in Figure~\ref{fig:fig1}(Mid-Bottom).

\begin{figure}[t]
    \centering
    \includegraphics[width=\linewidth]{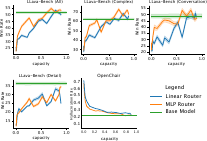}
    \caption{Evaluation of Elasti-LLaVA on LLava-Bench and OpenChair benchmarks against capacity. The capacity value indicates the percentage of image tokens selected by the \name routing module. Error bar for LLava-Bench results indicate 95\% confidence interval obtained via bootstrap sampling (100 samples with replacement).
    }
    \label{fig:llava}
\end{figure}

We evaluated Elasti-VLM's performance on both OpenChair \cite{benkish2024mitigating} hallucination benchmark and LLava-Bench \cite{llava_paper} open domain QA benchmark, and compared against performance of the base pretrained VLM model  (\texttt{llava-v1.5-7b}). We additionally experimented with more complex routing modules with an MLP router with 1 hidden layer and \texttt{GELU} activation function. As shown in Figure~\ref{fig:llava}, the LLava-Bench results show that Elasti-VLM is able to achive the same performance as the base model with 60\%~70\% of input tokens, and is able to even outperform the base model for complex and conversational tasks. However, for tasks that required detailed visual information such as LLava-Bench (detail) and OpenChair which evaluates VLM's ability to accurately describe all key objects in an image, Elasti-VLM offers no clear advantage over the base VLM model. 

When comparing routing modules of different complexities, results in Figure~\ref{fig:llava} suggests that the additional parameters in MLP routers is able to improve Elasti-VLM's performance, albeit at a significant computational cost (0.23\% additional routing parameters for MLP router compared to 0.00006\% for linear router as shown in Table~\ref{tab:param}). We leave further explorations of complexity of routing modules to \name performance for future research.

\section{Summary \& Discussion}
\name demonstrates that significant computational savings can be obtained through learned routing across modalities, and can be readily incorporated into existing post-training pipelines with techniques such as LoRA. Our method far out-performs static pruning of model parameters with minimal additional parameters and robust learned routing, which provides practical tools for not only model efficiency but also interpreting learned representations in various components of the transformer architecture. 

An interesting finding of the current work, and perhaps a key limitation, is that the computational savings of skipping tokens from processing by attention modules do not perform as well as other sources of computational savings (skipping attention heads, skipping MLP blocks, reducing MLP active parameters). While this performance degradation can be rescued by very low-rank LoRA adapters for the attention module, it raises the question of the type of redundancy that exists in attention computation. We hypothesize that the comparatively poor performance of input subset selection for MHA is due to the design of the routing module, which performs routing based on \emph{individual} token embedded in a way that is agnostic to context. Indeed, choosing the tokens that have the least impact on MHA module output requires knowing \textit{a priori} the degree to which this token will be attended to by other tokens. One potential improvement of the current formulation of \name is to compute the routing of tokens for the MHA module based on not only the token embedding but also the attention weights of previous layers. We leave explorations of alternative methods for routing tokens for MHA layer for future works.

\clearpage
\bibliographystyle{plain}
\bibliography{main}

\clearpage
\newpage
\appendix
\section{Pruning Experiments for Analyzing Redundancy in Pretrained LLMs}
\label{sec:pruning}
Here we describe the procedure in which the redundancy in pretrained LLMs is studied as reported in Section~\ref{sec:redundancy}.

\subsection{Model and Setup}\

We utilized the \textbf{Gemma-2-2b-it} model as the base pretrained LLM for our experiments. The redundancy was analyzed by systematically removing components of the model and evaluating the resulting impact on its performance. Specifically, we focused on two key architectural components:
\begin{itemize}
    \item \textbf{Entire Transformer Layers:} Skipping entire layers in the Transformer architecture.
    \item \textbf{Multi-Head Attention (MHA):} Removing individual attention heads within the Transformer layers.
\end{itemize}

\subsection{Component Removal Procedure}

\textbf{Random Selection:}
\begin{itemize}
    \item We progressively skipped a random subset of full Transformer layers or removed a random subset of attention heads within the MHA module. Random selection ensured that the skipping process did not introduce systematic bias.
    \item For each target number of components to remove (e.g., 3 Transformer layers or 3 attention heads), we randomly selected 5 distinct groups of components and performed removal separately. The final results for each configuration were obtained by averaging the performance metrics across these 5 groups.
    \item The number of skipped Transformer layers or removed attention heads was increased incrementally in each experimental configuration to evaluate performance degradation at different levels of model pruning.
\end{itemize}

\noindent
\textbf{No Additional Training:}

\begin{itemize}
    \item After skipping Transformer layers or removing attention heads, no additional learnable parameters were introduced. The remaining model weights were frozen, with no fine-tuning or retraining performed.
\end{itemize}

\subsection{Evaluation Metrics}
The impact of skipping Transformer layers or removing attention heads was quantified using the following metrics:
\begin{itemize}
    \item \textbf{Language Modeling Loss Difference ($\Delta$ LM Loss):}
    The difference in LM loss between the base model and the modified (pruned) model was computed as:
    \[
    \Delta \text{LM Loss} = \text{Loss}_{\text{pruned}} - \text{Loss}_{\text{original}}
    \]
    This metric captures the overall degradation in predictive performance due to skipping layers or components.
    \item \textbf{Top-1 Token Prediction Agreement (Top-1 Match):}
    The percentage of tokens for which the pruned model and the base model predicted the same vocabulary index. This is calculated as:
    \[
    \text{Top-1 Match} = \frac{\text{Count}_{\text{matched}}}{\text{Total}_{\text{tokens}}}
    \]
    where \(\text{Count}_{\text{matched}}\) is the number of tokens with identical predictions, and \(\text{Total}_{\text{tokens}}\) is the total number of tokens in the evaluation dataset.
\end{itemize}

\subsection{Experimental Tasks and Datasets}
The experiments were conducted on two datasets with distinct characteristics:
\begin{itemize}
    \item \textbf{GSM8K:} A benchmark for mathematical reasoning tasks.
    \item \textbf{HumanEval:} A benchmark for code generation tasks.
\end{itemize}

\subsection{Results Analysis}
The results, visualized in Figure 2 of the main text, reveal the following:
\begin{itemize}
    \item Skipping a small number of Transformer layers or attention heads caused minimal performance degradation across both datasets.
    \item \textbf{Skipping Transformer Layers:} Performance degradation was more severe as the number of skipped layers increased, with a sharp rise in LM loss and a significant reduction in Top-1 Match.
    \item \textbf{Removing Attention Heads:} The model exhibited greater tolerance to the removal of individual attention heads, with slower performance degradation compared to skipping entire Transformer layers.
    \item The differing rates of performance degradation between GSM8K and HumanEval suggest that redundancy is task-dependent.
\end{itemize}

\section{Implementation Details of \name}
\label{sec:implementation}
\name lowers model redundancy through two key mechanisms: input selection and parameter selection. Input selection determines which tokens (inputs) are processed by the transformer layer and which are skipped. Once a token passes through the input selection stage and enters the layer, parameter selection identifies the specific parameters to be utilized during computation. In this section, we will discuss the implementation of each mechanism in detail.
\subsection{Input Subset Selection}

\paragraph{Routing Module}
Given a set of tokens $X=\{x_1, x_2, ... x_T\} \in \mathbb{R}^D$ as input to a block $B: \mathbb{R}^D \rightarrow\mathbb{R}^D $, the routing module $R: \mathbb{R}^D \rightarrow [0,1]$ produces a set of routing scores $R(T)=\{s_1, s_2, ..., s_T\}$ which serve as the basis for token selection. The output of block $B$ with router $R$ is given by:  
\[
B_R(X)=\{B(x_t)\cdot \mathbbm{1}[R(x_t)>\theta]\cdot R(x_t)+x_t|x_t\in X\}
\label{eq:block_output}\]
Where $\mathbbm{1}: \mathbb{R} \rightarrow \{0,1\}$ is an indicator function that returns 1 if $R(x_t)>\theta$ and 0 otherwise. The parameter $\theta$ acts as the decision threshold for input selection.

The threshold for selection is decided by the capacity factor $c$, a hyperparameter. We select tokens with top $k$ router scores during training, with $k=c\cdot T$. However, this is not feasible during inference for causal models: due to the auto-regressive nature of causal language models, the distribution of the routing weights can change during the generation process, and thus the top $k$ tokens as well, therefore we use a decision threshold of 0.5 during inference instead.

The selection threshold $\theta$ is determined by a hyperparameter called the capacity factor, $c$. During training, tokens are selected based on the top $k$ routing scores, where $k=c\cdot T$. However, this approach is impractical during inference. Due to the auto-regressive nature of language models, the distribution of routing weights can vary throughout the generation process. Consequently, the set of top $k$ tokens may also change dynamically, making static selection infeasible. Therefore we, use a decision threshold of 0.5 during inference instead.

\paragraph{Auxiliary Loss}
During inference, the router does not enforce the capacity constraint used in training, as it relies on top $k$ selection. This discrepancy can lead to the router selecting significantly more tokens than intended, as the gradient produced by \eqref{eq:block_output} may encourage such behavior. To ensure consistent behavior during inference, we introduce an auxiliary loss to train the router to select a number of tokens close to the capacity factor. The auxiliary loss is defined as:
\[
\mathcal{L}_{\text{top-k}}=-{(y\log(R(X)) + (1 - y)\log(1 - R(X)))}
\]
Here, $y$ is a binary sequence of length $T$, with $c\cdot T$ entries set to 1 (being positives) and $(1-c)\cdot T$ of the entries set to 0 (negatives). The positively selected tokens correspond to those that are selected in the selection step and processed by $B$. This approach trains the router to maintain a selection count close to the capacity factor even during the inferencing stage.
\subsection{Parameter Subset Selection}
\paragraph{Routing Module}
Redundancy in transformer models can arise not only from input tokens but also from the model's parameters. To address this, we eliminate parameter redundancy by focusing on attention heads and MLP experts. Similar to input selection, for a block $B$, we define a router $R: \mathbb{R}^n \rightarrow [0,1]^M$ where $M$ is the number of experts represents the number of experts (i.e., individual attention heads or subdivisions of the MLP block). The router selects the top $k$ experts with the highest routing scores, where $k$ is a hyperparameter. Unselected experts simply pass their input through without modification, while the selected experts process their input as usual.

\paragraph{Auxiliary Loss}
The selection of experts can be influenced by the initialization of the routing module. Poor initialization may lead to certain experts being persistently selected while others remain idle, irrespective of the input. To mitigate this, we apply a load-balancing loss as described in \cref{sec:objective}: 
\[
\mathcal{L}_{\text{load}}=\sum^M_{m=1} \text{count}(\{m|m\in \text{top-k}(R(X))\}) \cdot R(X)_m
\label{eq:block_output}\]
Here $\text{count}(\{m|m\in \text{top-k}(R(X))\}$ calculates how often expert $m$ appears in the top-k selection by the router $R$. This loss penalizes experts with higher top-k counts by lowering their routing probabilities while boosting the probabilities of less frequently selected experts. As a result, it ensures more balanced expert utilization and prevents certain experts from remaining idle.

\section{\name Model Output Examples}
\subsection{Elasti-LLM}
\begin{figure*}[h]
    \centering
    \includegraphics[width=0.8\linewidth]{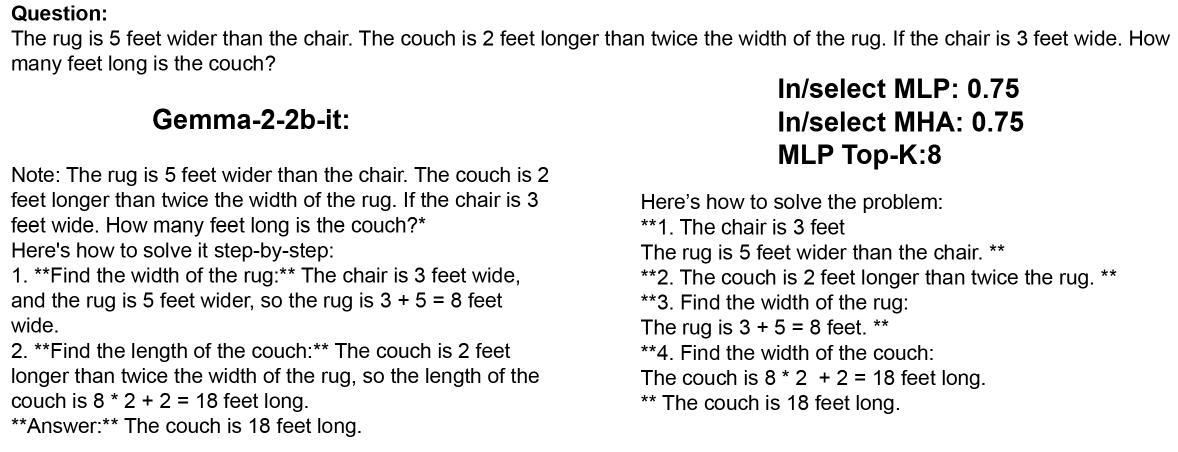}
    \caption{Example output of Gemma-2-2b-it. The capacity of input selection for MLP/MHA and parameter selection are listed above the outputs. The output quality, despite being influenced by the MLP and MHA capacity, still returns the correct answer and reasoning steps.}
    \label{fig:gemma_generate_results}
\end{figure*}

In figure \Cref{fig:gemma_generate_results}, we provide example outputs of pretrained \texttt{Gemma-2-2b-it} and its elastic counterpart with input subset selection for both MLP/MHA modules with a capacity factor of 0.75 and parameter subset selection for MLP module with capacity 8 (half of the experts).)
\subsection{Elasti-ViT}
\begin{figure}[h]
    \centering
    \includegraphics[width=\linewidth]{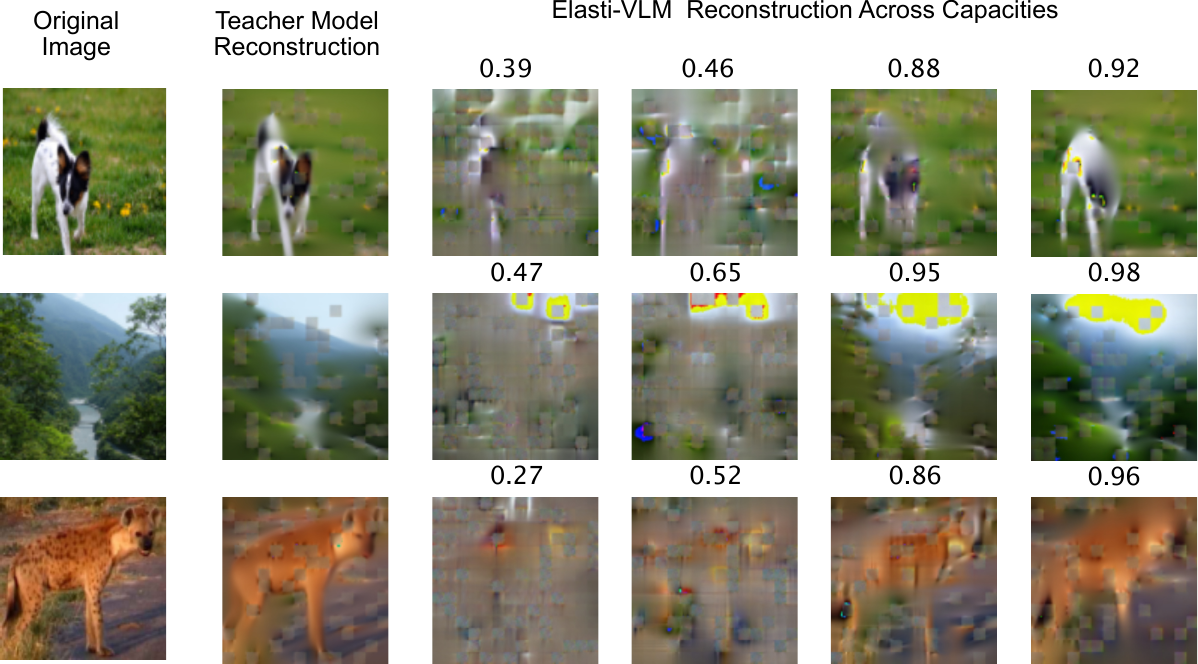}
    \caption{Example reconstruction (decoder output) of ImageNet images from pretrained ViT-MAE and Elasti-ViT with different capcaity factors. Each ELasti-ViT subfigure is labeled with cosine similarity against the pretraind ViT-MAE's decoder output.}
    \label{fig:vit_different_cossim}
\end{figure}

We provide in Figure~\ref{fig:vit_different_cossim} examples of reconstruction from the Elasti-ViT model across capacities. Each subfigure of the Elasti-ViT reconstruction is labeled by the cosine similarity between the decoder output of Elasti-ViT and the decoder output of the pretrained ViT-MAE model.

\subsection{Elasti-VLM}
\begin{figure*}[h]
    \centering
    \includegraphics[width=0.7\linewidth]{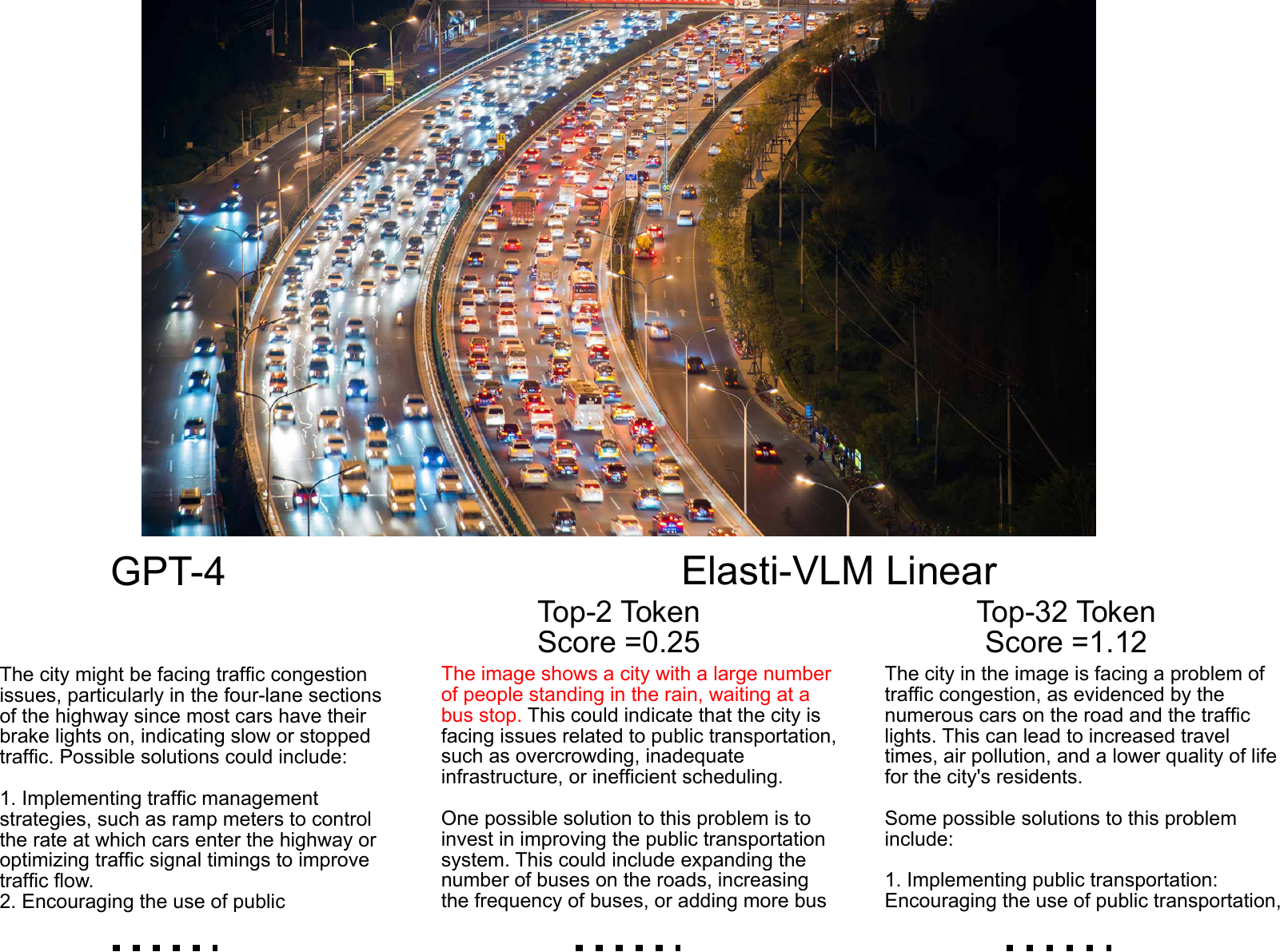}
    \includegraphics[width=0.7\linewidth]{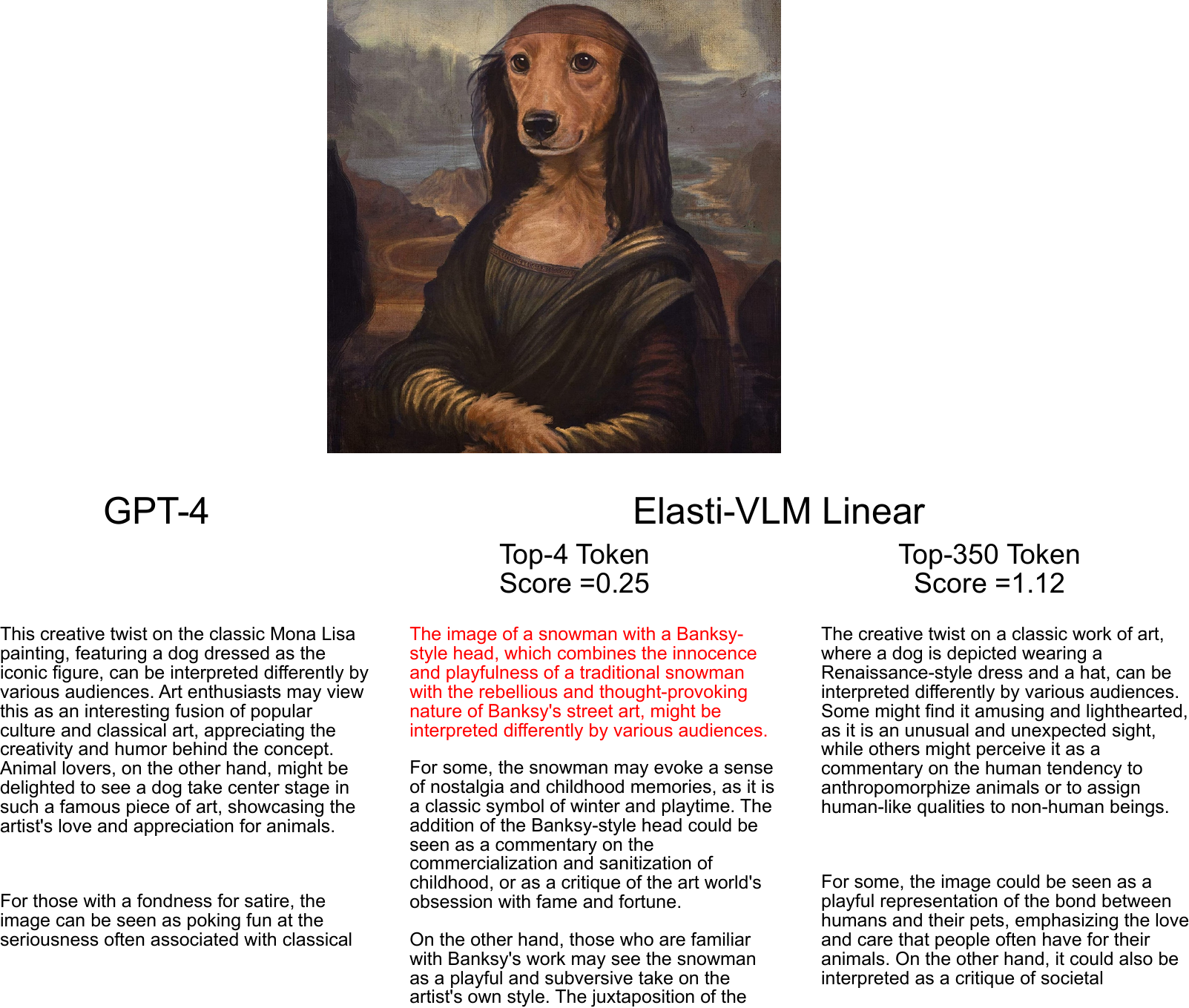}
    \caption{Example responses of LLava-Bench comparing GPT-4 response and Elasti-VLM response for different top-k image tokens. Errors made by Elasti-VLM are highlighted in red, which led to the low score (score=1 means Elasti-VLM performs as well as GPT-4).}
    \label{fig:vlm_examples}
\end{figure*}

We provide example outputs of pretrained \texttt{LLaVA-1.5-7B} and its elastic counterpart as shown in Figure~\ref{fig:vlm_examples}. According to LLaVA-Bench evaluation script, the scores were computed as the ratio between the score of the Elasti-VLM response and that of the GPT-4 response. The scores were given by \texttt{llama-3.1-70b-instruct} with temperature 0 using the same instruction templates as LLava-Bench.

\end{document}